\documentclass[letterpaper,journal]{IEEEtran}
\usepackage{amsmath,amsfonts}

\usepackage{array}
\usepackage[caption=false,font=normalsize,labelfont=sf,textfont=sf]{subfig}
\usepackage{textcomp}
\usepackage{stfloats}
\usepackage{url}
\usepackage{verbatim}
\usepackage{graphicx}
\usepackage{epstopdf}
\usepackage{cite}
\usepackage[linesnumbered,ruled,vlined]{algorithm2e}
\usepackage{multirow}
\usepackage{amssymb}

\newcommand{\Call}[1]{\textsc{#1}}

\begin{document}

\title{VersaDB: A High-Performance AI Storage Database for Unifying Mutimodal Datasets}

\author{
    Cong Wang\textsuperscript{$\dagger$}, Zelin Liu\textsuperscript{$\dagger$}, Yang Luo\textsuperscript{$\dagger$}, Ran Zhang, Zhijian Guo, Hui Zhang,Fan Yu, Yanfei Cao, Naijie Gu\textsuperscript{$\ddagger$}, Jun Yu\textsuperscript{$\ddagger$}
    \thanks{
        This work was supported by National Key R\&D Program of China under Grant No.2021ZD0110400
        
        This work was supported by the University of Science and Technology of China (First Affiliation) and the Huawei (Second Affiliation) MindSpore open source project.%

        $\dagger$ These authors contributed equally to this work.%
        
        $\ddagger$ These authors are co-corresponding authors.%
        
        Naijie Gu, Hui Zhang, Jun Yu, Yanfei Cao are with the University of Science and Technology of China, Anhui, China(e-mail:\{gunj,fzhh,harryjun\}@ustc.edu.cn, caoyf@mail.ustc.edu.cn).

        Ran Zhang is with the South China University of Technology, Guangzhou, China(e-mail: 202164010462@mail.scut.edu.cn).
        
        Zelin Liu is with the Shanghai Jiao Tong University, Shanghai, China(e-mail: julio1@sjtu.edu.cn).

        Cong Wang, Yang Luo, Fan Yu, and Zhijian Guo are with Huawei, China(e-mail:\{wangcong64, luoyang42, fan.yu, guozhijian\}@huawei.com).
    }
}

\markboth{Journal of \LaTeX\ Class Files,~Vol.~14, No.~8, August~2021}%
{Shell \MakeLowercase{\textit{et al.}}: A Sample Article Using IEEEtran.cls for IEEE Journals}

\IEEEpubid{0000--0000/00\$00.00~\copyright~2021 IEEE}

\maketitle

\begin{abstract}
The AI field has been rapidly developing, leading to the emergence of a large number of AI training datasets of various types. These datasets contain different modalities, including text, images, audio, etc., and may come in various data storage formats. With the advancement of AI hardware, AI computation units like GPUs, TPUs, and NPUs can greatly accelerate the training speed of AI models, which in turn increases the demand for faster data processing. When using existing AI processing frameworks to handle datasets with different modalities and storage formats, processing speeds may be suboptimal due to issues such as data layout and the way users handle the data. Therefore, using a unified database to store multiple data formats can better manage and optimize data access. In this paper, we introduce VersaDB, a database designed specifically for AI datasets with various modalities. We implemented a page-based storage system, separating structured and unstructured data. Additionally, we generated B+ tree-based index files to accelerate data access. VersaDB supports automatic sharding and maintains a hierarchical metadata management system, with corresponding metadata maintained at the page, shard, and global levels, forming the foundation for the efficient operation of the database. We also focused on ease of use by providing APIs for directly converting datasets into VersaDB, as well as APIs for converting popular AI data storage formats (e.g., CSV, TFRecord, .bin) into VersaDB.Our experiments show that using VersaDB can achieve up to 5.35x acceleration and maintain consistent performance across different parallelism levels.
\end{abstract}


\section{Introduction}
\IEEEPARstart{W}{ith} the rapid development of multi-sensor fusion applications and AI, we are closer than ever to creating an AI agent capable of perceiving the world and interacting in a human-like manner using multiple senses\cite{1}. AI advances are driven by big data\cite{26}, and recent datasets exhibit trends of richer modalities and larger scales\cite{24}, such as Conceptual 12m\cite{2} Laion-400m\cite{3} X-world\cite{4}. With the development of AI computing hardware, such as Nvidia's GPUs, Google's TPUs, Huawei's NPUs, and accelerators designed for specific networks\cite{37}\cite{38}, along with various model optimization techniques\cite{39}, data loading may become a bottleneck in AI training \cite{5}\cite{21}. This is not only because the improvement in computational power and efficiency increases data throughput but also because data publishers release AI data in different formats, researchers are unable to read the data using a unified and efficient data reading architecture, leading to significant performance variations between different datasets.

To accelerate data reading and processing, various AI frameworks use different methods. For example, Pytorch\cite{22}, Mindspore, and Tensorflow all support multi-threaded parallel processing, distributing data loading and preprocessing tasks across multiple threads to achieve parallel processing, thereby reducing waiting time during model training or inference. Prefetching is another effective method. It works by loading data in advance so that it is ready in memory when needed. This effectively overlaps GPU computation with CPU data reading time, reducing the impact of I/O time\cite{5}. However, acceleration techniques in AI frameworks face several issues: Firstly, increasing the number of threads can lead to additional overhead\cite{31} (e.g., thread switching and inter-process communication), and prefetching is only effective when model training time exceeds data processing time\cite{6}. Secondly, acceleration techniques typically cannot change the layout and organization of data, and lack in indexing systems, which may slow down data indexing or increase memory overhead. Thirdly, the issue of requiring different reading architectures for different modal data has yet to be solved, potentially leading to inconsistent performance.

Normalizing data to a unified format or database is a convenient approach for management and reading. Additionally, rearranging the data can lead to better reading performance. For example, the LMDB storage database stores key-value data using a B+ tree, offering features such as multi-version concurrency control, fast disk I/O, and others\cite{7}. Due to its ability to provide rapid data access\cite{30}, it is suitable for deep learning tasks. Similarly, converting data to TFRecord is another common choice\cite{27}\cite{28}\cite{29}. TFRecord\cite{8} is an efficient data storage format designed specifically for Tensorflow, aiming at improving the efficiency of loading and processing large-scale datasets. 
\IEEEpubidadjcol
Data stored in TFRecord are serialized in the protobuf binary format, supporting parallel reading and prefetching using Tensorflow APIs.

However, existing AI storage management solutions still have some issues. For example, due to the B+ tree structure of LMDB, it requires knowledge of the parent node and related information to locate the record's exact address. As a result, direct random access is not currently supported, and sequential access must be performed using a "cursor." TFRecord, designed specifically for Tensorflow, has a major issue of being incompatible with other AI frameworks. Furthermore, since both writing and reading in TFRecord are sequential, it cannot perform reading while writing. Both storage methods share some common issues: first, they do not support modifications to the data. If a user wishes to modify the data, the entire data file must be regenerated, which is very time-consuming when dealing with large datasets. Secondly, when storing data in memory, they do not distinguish between different modalities or data structures. LMDB uses key-value pairs to store all different data, and the data is generally stored in a serialized form. TFRecord, on the other hand, stores all different data sequentially, without considering the data types.

To address the above issues and the characteristics of AI data, we designed an AI database with the following considerations. First, our database should not only perform well during sequential reads but also handle random access efficiently, as shuffling plays a crucial role in AI training. It can significantly improve unbiased learning and convergence accuracy\cite{10}\cite{9}, and to some extent, prevent overfitting\cite{11}. Secondly, we considered the current organizational characteristics of AI data, such as the diversity of modalities (e.g., text, audio, video) that may be either structured or unstructured, with most corresponding labels being structured. We also considered the reading and writing of ultra-large datasets. Storing an entire large dataset in a single file is clearly unreasonable, as it would significantly impact data transfer and mobility.

We designed an efficient and user-friendly AI database called VersaDB. We divide the data storage into raw data and blob data sections for storing structured and unstructured data, respectively. In the data storage section, we implement page-based storage, with pages as the smallest storage unit, allowing for direct random access via page offsets. To accelerate data access, we create a separate index file outside the data storage file, implementing a B+ tree-based structure. We also employ a two-level sharding mechanism: the physical layer stores data, while the logical layer maps data through a distributed hash table. Shard information is managed by a hierarchical metadata system, which includes global, shard, and page-level dataset information, forming the foundation for the efficient operation of the data format. On top of these features, we support dynamic field extension capabilities and seamless integration of diverse data types, while also implementing a hierarchical lock manager to enable finer-grained locking, allowing data reading without affecting write operations.

\begin{figure}[t]  %
    \centering  %
    \includegraphics[width=0.48\textwidth]{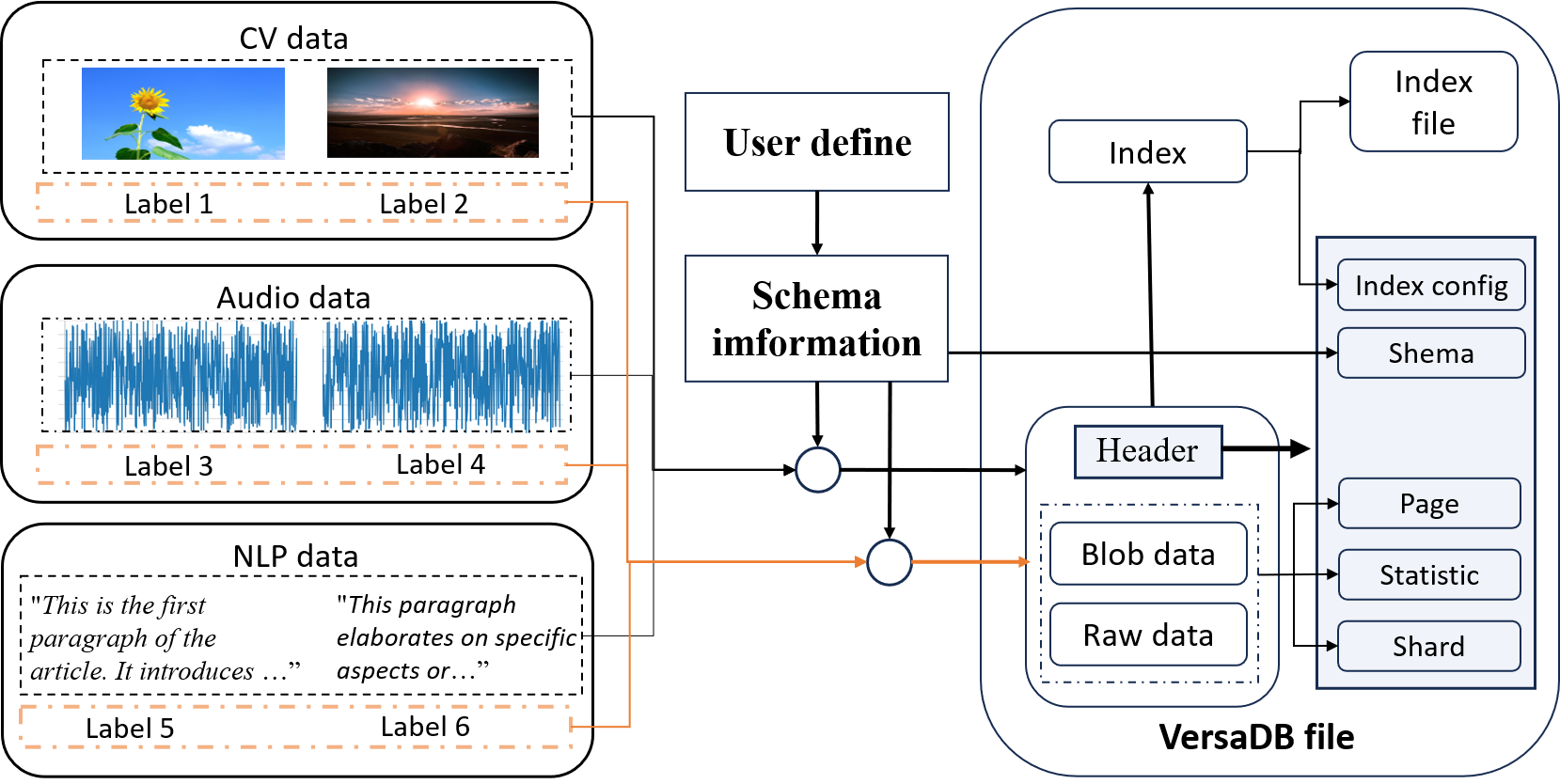}  
    \caption{VersaDB file generation workflow. VersaDB is designed to provide users with an efficient and convenient AI data storage management method. To store datasets of different modalities using VersaDB, users need to define the schema information based on the content to be stored. Then, different data types in the dataset will be stored separately in Blob data pages and Raw data pages. For example, data labels will be stored in the raw data pages. Subsequently, an index file will be generated based on the stored data, and corresponding metadata will be included in the header.}  %
    \label{fig:intro_view}  %
\end{figure}

Fig. \ref{fig:intro_view} illustrates the VersaDB file generation workflow. The process begins by parsing input data according to the user-defined schema, writing structured data to the raw data section and unstructured data to the blob data section. During this process, the system generates statistics and shard metadata while constructing the index structure. The final step integrates the header, raw data, and blob data into a cohesive data file.

In summary, this paper makes the following contributions:
\begin{itemize}
    \item \textbf{AI-specific adaptation.} The AI database we designed implements a page-based storage model, where data is stored separately according to its type. It can automatically generate multi-level index files based on the B+ tree structure according to the stored data, and meticulously maintain multi-dimensional metadata. This design fully adapts to the diverse ecosystem of current AI datasets and meets the demand for fast data access.
    \item \textbf{High performance.} We tested multiple datasets from common modalities such as images, text, and audio in two different machine environments. The test tasks included Pure reading, Shuffle reading, and Preprocessing, achieving up to 5.35x acceleration under default conditions. We also evaluated the performance of VersaDB at different parallelism levels and distributed levels. The experimental results demonstrate that our data structure offers exceptional performance and scalability.
    \item \textbf{Memory consumption Optimization.} The memory consumption of VersaDB has been specially optimized. We use the HyperLogLog structure for cardinality estimation, apply specific compression schemes to both the outer and inner layers of pages, and implement strategies such as the scalar-blob separation strategy and dynamic page replacement policy.
\end{itemize}

\section{Background}
The training of an AI model generally involves the following steps: data reading, data preprocessing, and model training. A batch of data (usually ranging from 32 to 512 samples) is first loaded into the host memory, and then parsed according to its data type. For example, images need to be parsed from storage formats like JPG or PNG into arrays or tensors, while audio may require parsing the WAV format to extract channel data. After parsing, preprocessing is usually performed before feeding the data into the AI model for training. Existing deep learning frameworks like TensorFlow, PyTorch, and Mindspore already cover the above model training process. Below, we will briefly introduce how these frameworks process data and some common AI data storage formats.

\textbf{AI training framework}.When training an AI model, it is often necessary to leverage AI training frameworks such as PyTorch, TensorFlow, or Mindspore. These frameworks provide comprehensive support for tasks ranging from data processing to network training, facilitating the building and training of AI models. Different frameworks define their own data processing workflows. For instance, PyTorch manages data through the Dataloader, TensorFlow defines its own tf.data API for data management, and Mindspore defines data processing as a graph, employing multi-stage parallel pipelines. AI data formats rely on AI training frameworks for reading, and the speed of data loading directly affects subsequent data preprocessing and model training. Therefore, an efficient AI data storage management method is essential to ensure a fast end-to-end AI training pipeline.

\textbf{AI data storage formats}. Current AI datasets come in various storage formats. For example, the JSON format is easy to parse and generate, supports multiple data types, and allows for nested structures\cite{23}. It is commonly used to store image information, category data, annotations, and other structured data. BIN files are a type of binary data storage format that can occupy less space and store more complex data structures, such as serialized images or audio. However, BIN files lack readability, do not include structural descriptions of the data, and may require external documentation for proper parsing. The CSV format is a widely used text-based format with high compatibility and readability. However, it does not support nested data structures and lacks the capability to handle large-scale datasets efficiently.

\textbf{AI distributed training}. Model parallelism and data parallelism are commonly used strategies for training large models today\cite{40}. Model parallelism involves placing different parts of the model on different compute cards. In the data parallelism strategy, the dataset is split into multiple chunks, and different compute cards read different data chunks to accelerate the reading of ultra-large datasets.

\begin{figure*}[t]  %
    \centering  %
    \includegraphics[width=0.85\textwidth]{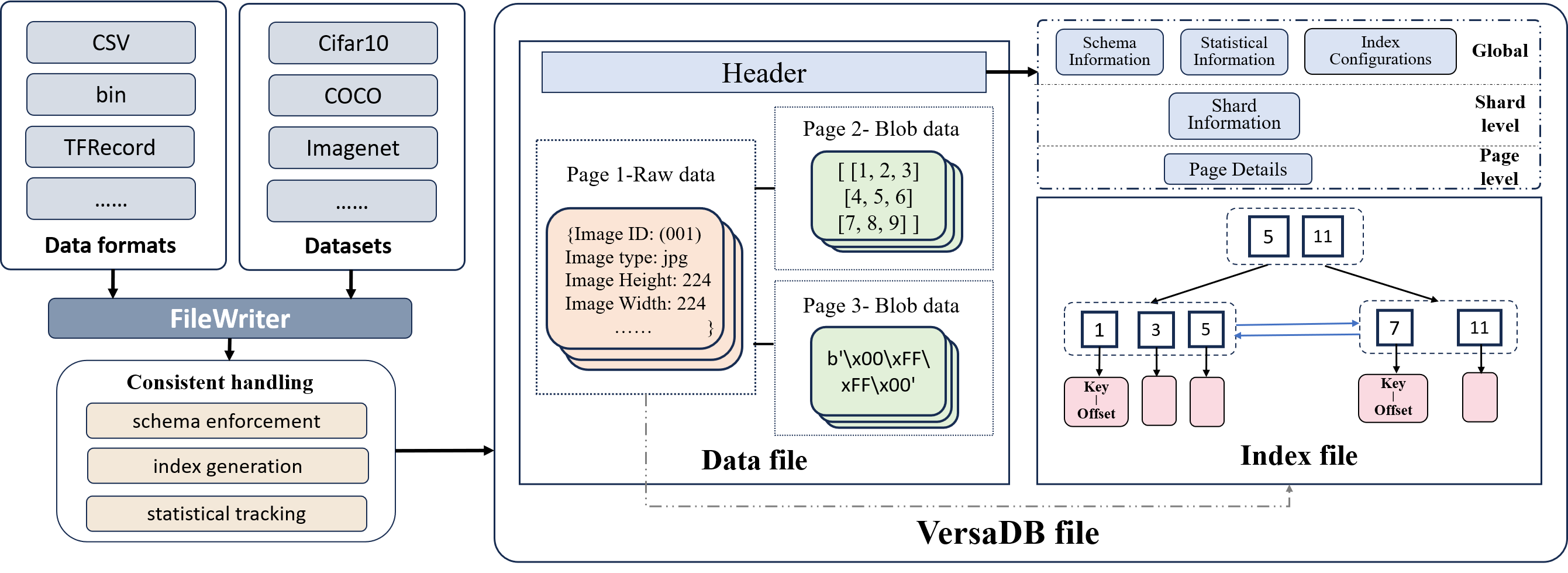}  
    \caption{This paper introduces the architecture of VersaDB. VersaDB is divided into two main components: the Data file and the Index file. The Data file is used to store the actual data and metadata, while the Index file is used to assist in locating the physical storage location of the data. In this paper, we will primarily focus on the data storage methods, the APIs required to interact with VersaDB, the statistics and structure of metadata, data sharding management, dynamic page management, as well as random sampling and parallel reading optimizations.}  %
    \label{fig:overview}  %
\end{figure*}
\textbf{Multimodal AI training}. Training AI with multimodal methods, which involves extracting and associating information from diverse types of data \cite{32}, offers great promise and potential across a variety of applications. For instance, research has shown that incorporating two or three modalities in tasks such as sentiment analysis can significantly improve effectiveness\cite{33}\cite{34}. The training of multimodal AI models generally follows two main strategies: feature-level fusion\cite{35} and decision-level fusion\cite{36}. In feature-level fusion, data from multiple modalities is combined into a unified feature vector, enabling the integration of these modalities. In contrast, decision-level fusion allows independent models to process each modality separately, produce individual decisions, and then combine these decisions to associate the multimodal information. Both fusion strategies, whether at the feature or decision level, are impacted by the fact that the slowest modality during training can act as a bottleneck, restricting the performance of the entire model. Moreover, the performance of datasets with varying storage structures or modalities can differ during data retrieval. As a result, ensuring efficient processing of each modality is essential for optimizing model performance.

\section{VersaDB overview}
VersaDB implements a hierarchical data management system
optimized for large-scale, multimodal datasets through a
comprehensive technical framework, as shown in Fig. \ref{fig:overview}. This section details the
core components and their technical implementations:

In section IV, we present the hierarchical file structure, which implements a binary file structure combining a data file and an index
file. The data file employs a B+ tree-based organization with
a header containing critical metadata, a raw data section for
structured data storage, and a blob data section for unstructured
content. This implementation enables efficient data access
through direct page addressing and optimized index lookups.

In section V, we describes the implementation
of three core interfaces: FileWriter, FileReader, and VersaPage. These interfaces utilize buffered I/O mechanisms, multithreaded access patterns, and LRU-based page caching to
optimize data operations. Specialized conversion modules
implement format-specific optimizations while maintaining
consistent schema validation and write orchestration.

In section VI, we detail the
system implementation for schema definition and the
hierarchical metadata organization. The schema system employs
a binary encoding format with version control mechanisms,
while metadata management implements a two-phase commit
protocol with write-ahead logging for consistency guarantees.

The section VII
presents the two-level partitioning mechanism for shards,
the distributed aggregation tree for statistics, and the hybrid
indexing approach combining B+ trees and inverted indexes.
These components implement specific algorithms for distributed
consistency and efficient query processing.

Section VIII implements a dynamic storage unit
system with hybrid compression mechanisms and buffer
pool management. The implementation utilizes a slot-based
architecture and copy-on-write mechanisms for concurrent
access handling.

Finally, the section IX details the
implementation of high-performance operators and parallel
execution mechanisms. The sampling and shuffling operators
employ modified reservoir and Fisher-Yates algorithms, while
parallelization support implements work-stealing task scheduling and multi-version timestamp ordering for consistency
control.

\section{hierarchical file structure}
VersaDB employs a hierarchical file structure consisting of a data file and an index file.The data file integrates a header section with metadata, a raw
data section for structured data storage, and a blob data section
for unstructured data, as illustrated in Fig. \ref{fig:overview}.

The header section maintains critical structural information
through five interconnected components. Schema Information
defines data types and their structural relationships, while
Shard Information maps logical shards to physical storage
locations. Page Details maintains the page table structure
and its relationships with shards, complemented by Statistical
Information that tracks multilevel metrics. Index Configurations
store the mapping between logical records and their physical
locations.

Within the raw data section, VersaDB implements pagebased storage, where each page serves as the minimal storage unit with a unique page ID mapped through Page Details.
The data within pages follows the schema-defined format,
enabling direct random access through page offsets. The blob
data section adopts a stream-oriented structure, utilizing an
offset table in the header for efficient blob localization. Each
blob entry contains a size field followed by the actual data,
optimizing variable-length object storage.

The index file implements a B+ tree structure to facilitate
efficient data retrieval. Leaf nodes store key-offset pairs
mapping index field values to physical locations in the data
file, while internal nodes maintain separator keys to accelerate
search operations. The leaf nodes are interconnected through
bidirectional links, enabling efficient range query processing.

This architecture achieves efficient data organization and
access through the separation of structured and unstructured
data storage, implementation of multilevel indexing, and
maintenance of fine-grained metadata. The B+ tree index
structure ensures logarithmic time complexity for data retrieval
operations, while the page-based storage system optimizes both
sequential and random access patterns.
\begin{figure}[t]  %
    \centering  %
    \includegraphics[width=0.5\textwidth]{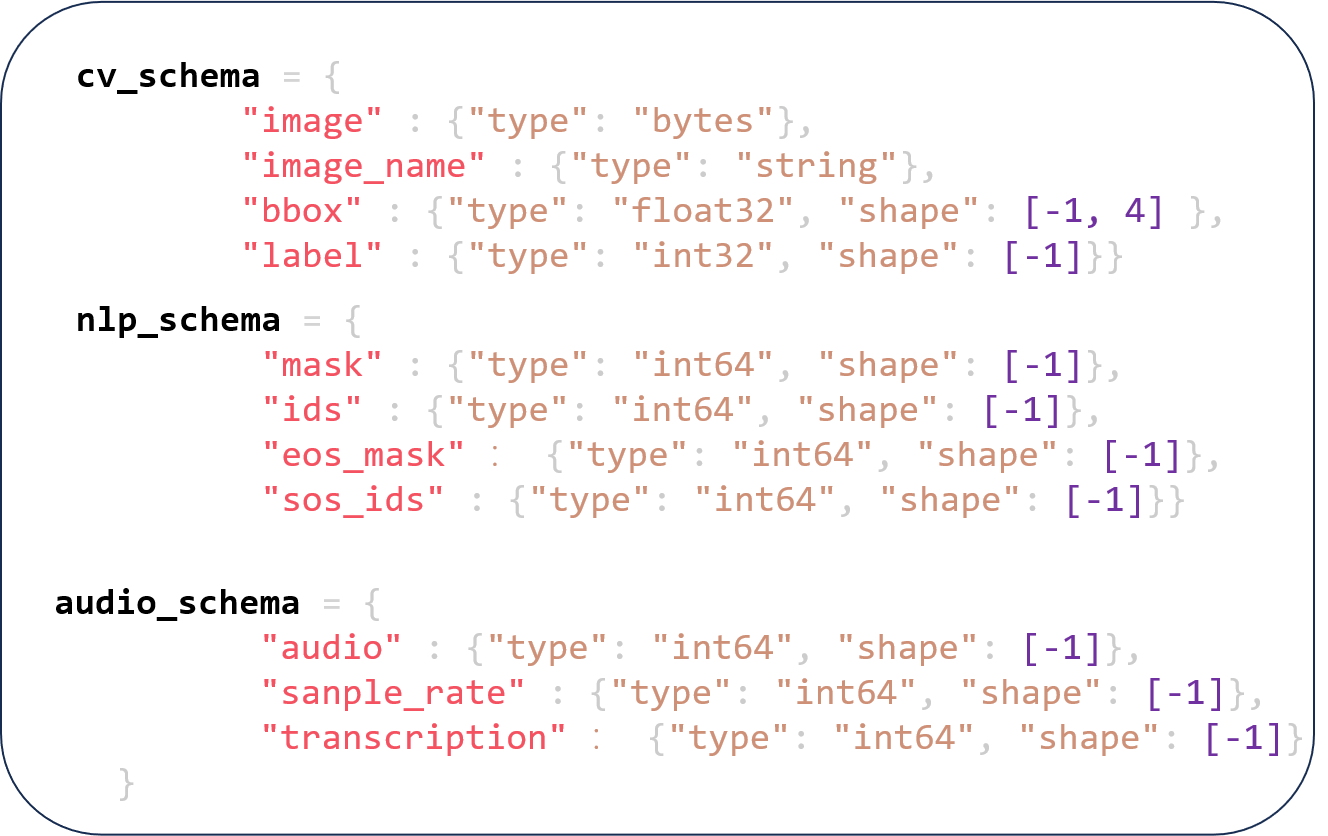}  
    \caption{Three simple definition for Schema  format.}  %
    \label{fig:shame}  %
\end{figure}

\section{User Interfaces}
VersaDB implements three core
interfaces: FileWriter, FileReader, and VersaPage, to enable
efficient data manipulation and access. The FileWriter interface
provides low-level write operations through a schema-driven
API. It implements a buffered writing mechanism where data
is first accumulated in memory buffers, then written to disk
in optimized chunks when the buffer reaches a predefined
threshold. This approach minimizes I/O operations while
maintaining memory efficiency.

The FileReader interface implements a multi-threaded reading mechanism that supports both sequential and random access
patterns. It utilizes a prefetching algorithm that predicts future
data access based on historical patterns and loads data into a
circular buffer. The interface exposes methods for field-selective
reading, allowing applications to retrieve only necessary fields
and reduce I/O overhead. The underlying implementation
maintains a thread pool that asynchronously loads pages while
the main thread processes the current data chunk.

The VersaPage interface extends the basic reading capabilities
by implementing a paginated access mechanism. It maintains
a page cache that stores frequently accessed pages using a
Least Recently Used (LRU) replacement policy. The interface
provides atomic operations for page-level access and updates,
ensuring data consistency in multi-threaded environments.

For dataset conversion, VersaDB provides specialized
conversion modules. These modules
implement dataset-specific parsing logic while sharing a
common write pipeline. For instance, the ImageNetToMR
module implements parallel image decoding and compression,
maintaining a producer-consumer pattern where decoder threads
feed into a shared write buffer. Similarly, the TFRecordToMR
module implements a streaming parser that directly maps
TFRecord protocol buffers to VersaDB schema definitions,
minimizing intermediate data copies.
Each conversion module inherits from a base converter
class that implements the core schema validation and write
orchestration logic. This design ensures consistent handling of
schema enforcement, index generation, and statistical tracking
across different dataset formats while allowing format-specific
optimizations in the parsing layer.

\section{Schema and Metadata Management}
\subsection{ Schema Management}
VersaDB implements a
hierarchical schema system that encodes data structure specifications through a type system. As shown in Fig. \ref{fig:shame}, each
schema definition consists of field descriptors that specify the
name, data type, and shape of each field. The type system
supports primitive types (integers, floats, strings) and complex
types (tensors, nested structures) through a composable type
hierarchy.

The schema implementation utilizes a binary encoding
format where each field descriptor is serialized as a sequence
of type, name, shape triplets. This format enables efficient
schema validation during data operations while minimizing
storage overhead. For nested structures, the system implements
a depth-first traversal mechanism that recursively validates
and processes nested fields according to their hierarchical
relationships.

Schema evolution is handled through a versioning mechanism
that maintains backward compatibility. When a schema update
occurs, the system generates a schema diff that captures field
additions, deletions, and modifications. These changes are
encoded in a version table within the header, allowing the
system to apply appropriate transformation rules when reading
data written with older schemas. The version table implements a
directed acyclic graph (DAG) structure to track schema lineage
and enable efficient schema resolution during data access.

\subsection{Metadata Management}
The metadata system in
VersaDB organizes dataset information in a hierarchical
structure across global, shard, and page levels. At each level,
metadata entries are encoded using a compact binary format
that optimizes both storage efficiency and access speed. The
system implements a lazy loading mechanism where metadata
is loaded on demand and cached in memory using a least recently-used (LRU) policy.

Global metadata maintains dataset-wide information including schema versions, statistical summaries, and index
configurations. Shard-level metadata implements a mapping
between logical shards and their physical storage locations,
along with shard-specific statistics and page allocation tables.
Page-level metadata tracks record boundaries, data distributions,
and bloom filters for efficient query optimization.

The metadata update process follows a two-phase commit
protocol to ensure consistency across distributed environments.
Updates are first written to a write-ahead log (WAL), then
applied to the in-memory metadata cache before being persisted
to disk. This approach guarantees atomic metadata updates
while providing crash recovery capabilities.

Statistical tracking is implemented through a hierarchical
counter system that maintains counts and distributions at
multiple granularities. Counter updates are performed using
atomic operations to ensure consistency in concurrent scenarios.
The system implements efficient aggregation algorithms that
can compute global statistics from shard-level counters without
requiring full dataset scans.

\section{Shard, Statistic, and Index Management}
\subsection{Shard Management}
VersaDB implements sharding through a two-level partitioning mechanism. At the physical
level, data is divided into fixed-size shards, each maintaining
its own page tables and local metadata. At the logical level,
a distributed hash table maps record keys to physical shards,
enabling efficient data distribution and load balancing. Each shard contains multiple pages organized
as the minimal storage units, with pages inheriting the parent
shard’s access control and consistency protocols.

The system implements atomic shard operations through
a distributed lock manager. Each shard maintains a version
counter and a write-ahead log for crash recovery. During
write operations, the system acquires shard-level locks before
updating both data and metadata, ensuring consistency in
concurrent scenarios. Read operations utilize a multi-version
concurrency control (MVCC) mechanism that allows concurrent
reads without blocking write operations.

\begin{figure}[t]  %
    \centering  %
    \includegraphics[width=0.45\textwidth]{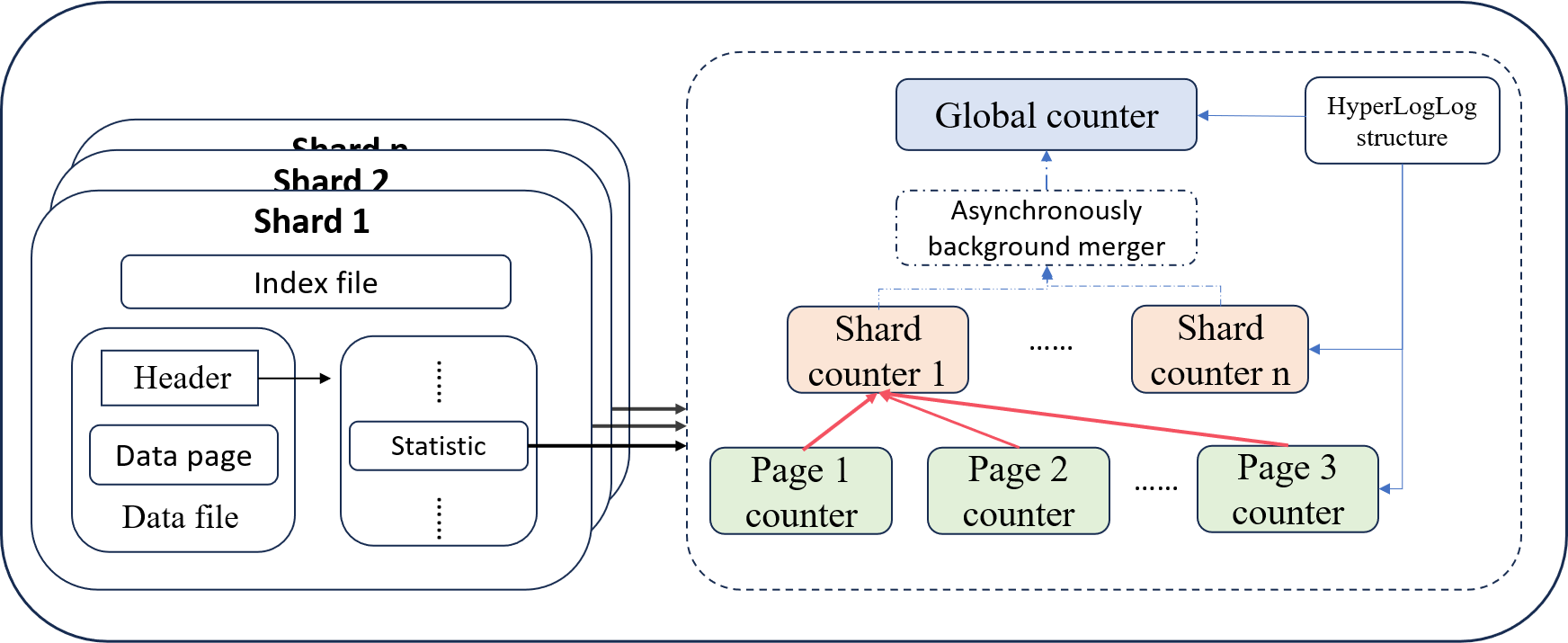}  
    \caption{Statistical counting based on distributed aggregation tree}  %
    \label{fig:statistic}  %
\end{figure}
\subsection{ Statistic Management}
The statistical framework maintains hierarchical counters at global, shard, and page levels
through a distributed aggregation tree, as shown in Fig. \ref{fig:statistic}. Statistics are generated during file creation and continuously
updated during write operations. The system implements approximate counting algorithms using HyperLogLog structures
for cardinality estimation, enabling efficient memory usage
while maintaining acceptable error bounds.

Counter updates follow an eventual consistency model where
local updates are immediately reflected in shard-level statistics while global aggregations are performed asynchronously.
The system employs a background merger that periodically
consolidates shard-level statistics into global counters using a
hierarchical aggregation protocol. This design trades immediate
global consistency for improved write performance while
ensuring that statistical queries eventually reflect the complete
dataset state.

\subsection{Index Management}
The indexing system implements
a hybrid approach combining B+ trees for structured data
and inverted indexes for unstructured content. The
system first builds primary indexes for raw data fields, then
constructs secondary indexes for blob data through parallel
processing. The Query procedure demonstrates the system’s
ability to handle both range-based and exact-match queries by
leveraging field-specific index structures.

Index updates are performed through a log-structured merge
(LSM) tree\cite{25} implementation. New entries are first accumulated
in an in-memory buffer, then merged into disk-based index
structures through background compaction processes. This
approach optimizes write performance while maintaining
query efficiency through careful compaction scheduling. The
system implements bloom filters at each index level to reduce
unnecessary disk accesses during query processing.

For distributed operations, the index manager maintains consistent views across nodes through a vector clock mechanism.
Each index update is tagged with a vector timestamp, allowing
the system to detect and resolve conflicts during concurrent
modifications. The index structures support efficient range partitioning, enabling parallel query processing across multiple
nodes while maintaining global consistency.

\begin{algorithm}
\caption{Multi-Strategy Sampling}
\label{Multi-Strategy}

\tcp{Category balanced sampling implementation.}
\SetKwFunction{FCategory}{CATEGORYBALANCEDSAMPLE}
\SetKwProg{Fn}{Function}{:}{}
\Fn{\FCategory{$data$, $config$}}{
    balanced\_samples $\gets$ [] \;
    categories $\gets$ \Call{GETCATEGORIES}(data) \;
    sample\_per\_category $\gets$ \texttt{config.total\_samples} / \texttt{categories.length} \;
    \For{each category in categories}{
        category\_data $\gets$ \Call{GETCATEGORYDATA}(data, category) \;
        samples $\gets$ \Call{RANDOMSAMPLE}(category\_data, sample\_per\_category) \;
        balanced\_samples.\Call{EXTEND}(samples) \;
    }
    \Return balanced\_samples \;
}
\end{algorithm}
\section{ Page Management}
VersaDB implements a dynamic
page management system that serves as the fundamental storage
unit within shards. Each page implements a hybrid storage
format that combines a fixed-size header with variable-length
data segments. The page header contains a bitmap indicating
record validity, a checksum for data integrity verification, and
metadata describing the page’s internal organization. The data
segment employs a slot-based architecture where each slot
maintains offset pointers to record locations, enabling efficient
record insertion and deletion without page reorganization.

\begin{algorithm}
\caption{Category-Aware Shuffle}
\label{shuffle}
\tcp{Shuffling tasks while maintaining category alignment.}
\SetKwFunction{FMain}{CATEGORYSHUFFLE}
\SetKwProg{Fn}{Function}{:}{}
\Fn{\FMain{$tasks$}}{
    \tcp{Calculate the number of samples for each category.}
    category\_sizes $\gets$ \Call{COUNTBYCATEGORY}(tasks)\;

    \tcp{Create a separate random sequence for each category.}
    category\_indices $\gets$ [ ] \;
    \For{each category in categories}{
        size $\gets$ category\_sizes[category] \;
        indices $\gets$ \Call{GENERATERANDOMSEQUENCE}(size) \;
        category\_indices[category] $\gets$ indices \;
    }

    \tcp{Reconstruct the sequence while maintaining category alignment.}
    aligned\_tasks $\gets$ [ ] \;
    \For{$i \gets 0$ \KwTo max\_category\_size}{
        \For{each category in categories}{
            \If{$i < $ category\_sizes[category]}{
                index $\gets$ category\_indices[category][i] \;
                task $\gets$ tasks[category][index] \;
                aligned\_tasks.\Call{APPEND}(task) \;
            }
        }
    }
    \Return aligned\_tasks \;
}
\end{algorithm}

The system implements page-level compression through a
dual-layer mechanism. The outer layer applies compression at
the page level using LZ4 for fast compression/decompression,
while the inner layer implements column-specific compression
schemes based on data characteristics. This hybrid approach
optimizes both storage efficiency and access speed by balancing
compression ratio with computational overhead.

Pages are organized within shards
using a B+ tree structure where leaf nodes contain page
metadata and pointers to physical page locations. The system
implements a buffer pool that caches frequently accessed pages
using a clock-based replacement algorithm. Page eviction
follows a dirty-page-first policy where modified pages are
persisted to disk before clean pages during memory pressure
situations.

Page splits and merges are handled through a copy-on-write
mechanism that maintains consistency during concurrent access.
When a page exceeds its size threshold, the system creates
a new page and redistributes records while maintaining a
linking structure for sequential access. This approach enables
dynamic page size adjustment while preserving data locality
and minimizing fragmentation.

For structured data access, the page manager implements row-group-based storage within pages, where records with similar
attributes are clustered together. This organization optimizes
scan operations by reducing I/O overhead during column-specific queries. The system maintains page-level statistics
including record count distributions and value ranges, enabling
the query optimizer to make informed decisions about access
patterns.
\begin{figure}[t]  %
    \centering  %
    \includegraphics[width=0.4\textwidth]{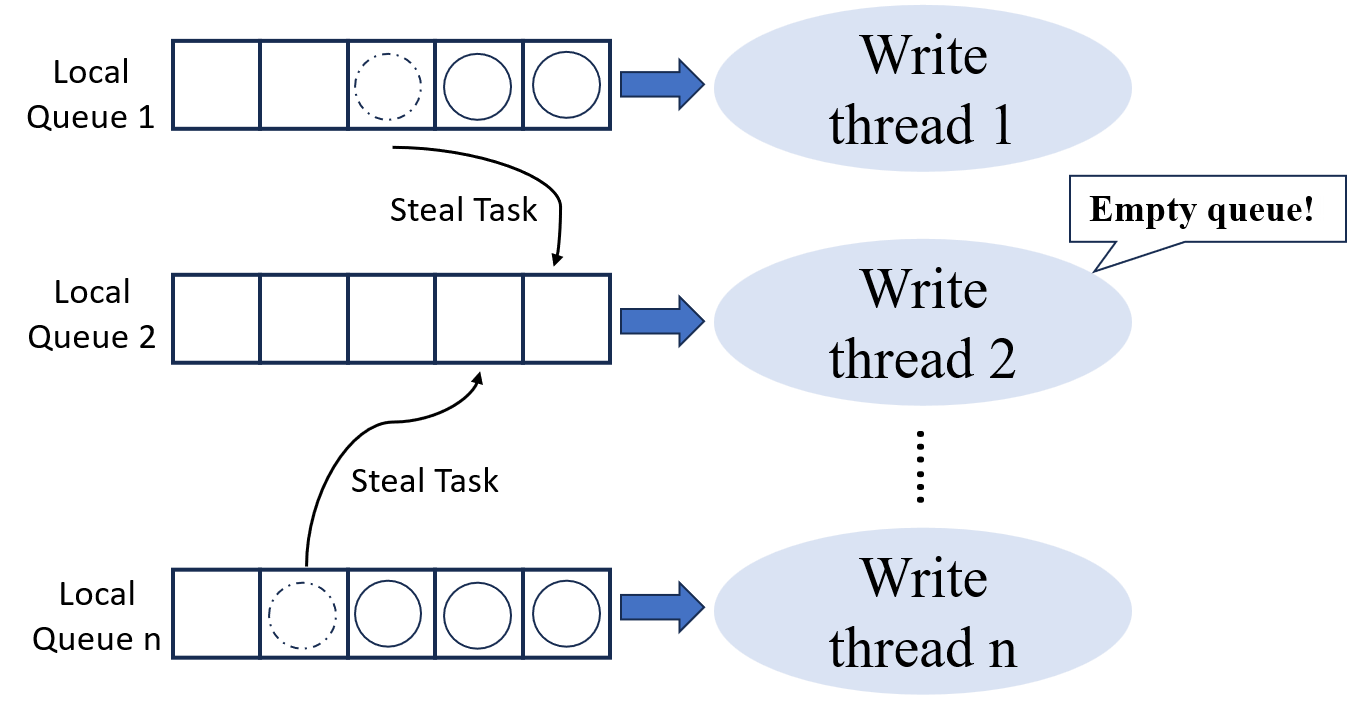}  
    \caption{When the local work queue of write thread 2 is empty, it will steal tasks from other queues to ensure load balancing.}  %
    \label{fig:steal_task}  %
\end{figure}

\begin{table}[b]
    
    \centering
    \caption{Number of Samples for Different Datasets in Various Domains}
    \renewcommand{\arraystretch}{1.5} 
    \begin{tabular}{c c c}
        \hline
        \textbf{Domain} & \textbf{Dataset} & \textbf{Number of Samples} \\
        \hline
        \multirow{4}{*}{CV} & VOC & 17,152 \\
                             & Kitti & 7,481 \\
                             & COCO & 118,287 \\
                             & ImageNet & 20,000 \\
                            
        \hline
        \multirow{3}{*}{NLP} & WikiText-103 & 100,000 \\
                             & SQuAD2.0 & 130,319 \\
                              & The Pile & 420,000\\
        \hline
        \multirow{2}{*}{Audio} & LibriTTS & 2174 \\
                               & Ljspeech & 2,426 \\
        \hline
        \multirow{1}{*}{Multimodal} & laion2B-300K & 300,000 \\
        \hline                     
        \multirow{1}{*}{Video}  & Youtube8M & 1,000,000\\
        \hline
    \end{tabular}
\end{table}

\section{Performance Optimization Modules}
\subsection{High-Performance Operators}
VersaDB implements optimized sampling and shuffling operators through
a multi-layered execution engine. The sampling system utilizes
a reservoir sampling algorithm with weighted selection probabilities, as detailed in Algorithm \ref{Multi-Strategy}. The algorithm maintains a
fixed-size reservoir and updates selection probabilities based on
category distributions, achieving O(1) space complexity while
ensuring representative sampling across categories.

The shuffle operator implements a distributed Fisher-Yates
algorithm modified for category awareness, as shown in
Algorithm \ref{shuffle}. The system first partitions records by category into
separate buffers, then applies independent random permutations
to each buffer. The final sequence is reconstructed through a
merge phase that interleaves samples from different categories
while maintaining their relative proportions. This approach
achieves O(n) time complexity and ensures uniform distribution
across categories.

For execution optimization, both operators implement a
pipeline architecture that overlaps I/O operations with computation. The pipeline consists of prefetch, process, and write
stages connected through lock-free queues. The prefetch stage
implements adaptive read-ahead based on access patterns, while
the process stage utilizes SIMD instructions for parallel data
transformation when applicable.

\begin{table}[h]
    \label{Experiment}
    \centering
    \caption{Experiment Setting}
    \renewcommand{\arraystretch}{1.5} 
    \begin{tabular}{p{1.5cm} p{5cm}} 
        \hline
        \textbf{Experiment} & \textbf{Categories} \\
        \hline
        \multirow{14}{=}{\parbox[t]{1.5cm}{Hardware \\ Configuration}} & 
        \parbox[t]{5cm}{\textbf{Machine 1}\\
        \hspace*{0.25cm}CPU INFO: Architecture: x86-64, 64bits+64cores, Clock Speed: 3.9GHz\\ \hspace*{0.25cm}DRAM INFO: Total memory size: 251Gi,Configured Memory Speed: 2934 MT/s\\ \hspace*{0.25cm}Framework Versions: PyTorch Version 2.4.0,TensorFlow version 2.13.1,MindSpore Version 2.2.14.}

        \\
        & \parbox[t]{5cm}{\textbf{Machine 2}\\ 
        \hspace*{0.25cm}CPU INFO: Model: Kunpeng 920, \\ Architecture: aarch64, 64bits+192cores, Clock Speed: 2.6GHz\\ \hspace*{0.25cm}DRAM INFO: Total Memory Size: 755Gi,Configured Memory Speed: 2933 MT/s  \\ \hspace*{0.25cm}Framework Versions: PyTorch Version 2.4.1,TensorFlow version 2.12.0,MindSpore Version 2.4.0.}\\ \\
        
        \hline
       
    \end{tabular}
\end{table}
\subsection{Parallelization Support}
The parallel execution engine
implements a task scheduling system based on a work-stealing
algorithm. As shown in Fig. \ref{fig:steal_task}, parallel write operations
are coordinated through a distributed task queue where each
writer thread maintains its local task queue. When a thread
exhausts its local queue, it attempts to steal tasks from
other threads’ queues, achieving dynamic load balancing with
minimal synchronization overhead.
The system implements fine-grained locking through a
hierarchical lock manager. Locks are acquired at the shard
level for write operations and at the page level for read
operations, enabling concurrent access to different pages within
the same shard. The lock manager uses a timeout-based
deadlock prevention mechanism where lock requests are aborted
if they cannot be satisfied within a specified time window.

Data consistency during parallel operations is maintained
through a multi-version timestamp ordering protocol. Each
write operation is assigned a timestamp, and conflicts are
detected by comparing operation timestamps. The system
implements a write-ahead logging mechanism where log
records are grouped by shard and flushed to disk using group
commit to optimize I/O performance while ensuring durability.

\section{EVALUATION}
\subsection{Experimental setup}
In this study, we evaluate three AI data management methods, each read by a corresponding AI training framework: LMDB is read by the PyTorch framework, TFRecord is read by the TensorFlow framework, and VersaDB is read by the MindSpore framework. We test these three data structures designed for AI training and compare them with raw dataset loading. The evaluation metrics include data sequential access speed, random access time, data preprocessing time, and resource utilization.

\textbf{System Configurations}: The experiments were conducted
on two systems with distinct hardware configurations, as summarized in Table II, to ensure comprehensive
evaluation.

\textbf{Datasets}:We evaluated the AI data management methods using various datasets spanning computer vision (CV), natural
language processing (NLP), video, mutimodal and audio domains. The data volumes tested in these datasets are shown in Table I. Among them, Imagenet, Wikitext\cite{16}, The Pile, Laion, and Youtube8M are all subsets of their respective total datasets.

\subsection{Performance analysis}
\begin{figure*}[t]  %
    \centering  %
    \includegraphics[width=0.9\textwidth ]{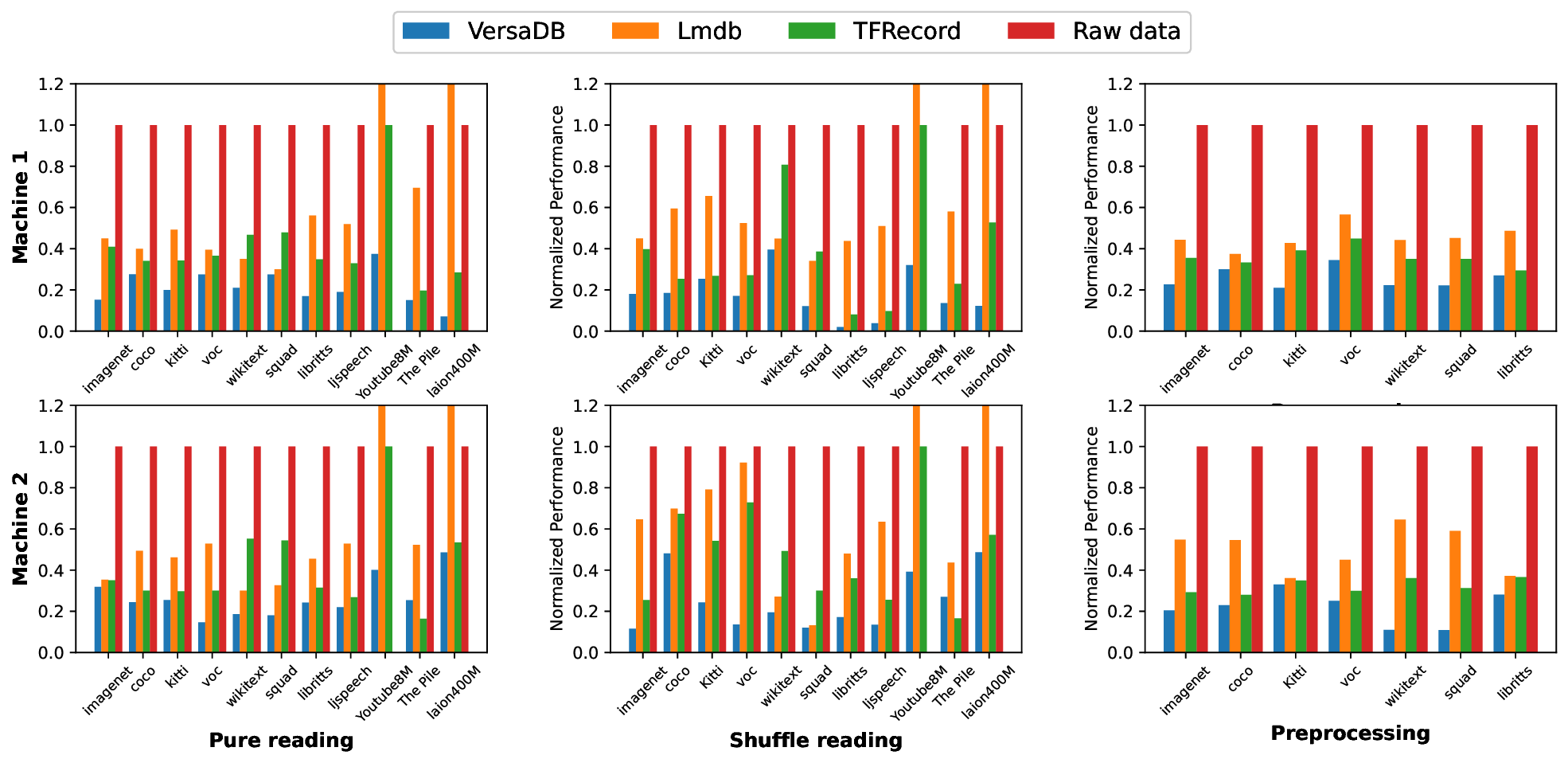}  
    \caption{Performance comparation of three different work}  %
    \label{fig:common datasets}  %
\end{figure*}

In this experiment, we tested the performance of VersaDB, Lmdb, TFRecord, and Raw data on eleven datasets, evaluating their performance mainly in three tasks: Pure reading, Shuffle reading, and Preprocessing, across two machine environments, as shown in Fig. \ref{fig:common datasets}. At the same time, for the three datasets Laion, Youtube8M, and The Pile\cite{20}, we did not test preprocessing, as these large-scale datasets are typically preprocessed and dimensionality-reduced for ease of transmission and use. For example, the images in Laion have already undergone feature extraction and are converted into 128-dimensional tensors. The raw format of the Youtube8M dataset is in TFRecord format. It is also important to note that, unless otherwise specified, the speedup comparison is made with the method that perform better between Lmdb and TFRecord.

\textbf{Machine 1}.  In the Pure reading task, the speedup of VersaDB compared to Lmdb and TFRecord ranges from 1.2x on the COCO\cite{14} dataset to 3.9x on the Laion\cite{3} dataset. In the Shuffle reading task, VersaDB shows a speedup from 1.06x on the Kitti dataset to 4.3x on the Laion dataset. In the Preprocessing task, the speedup of VersaDB ranges from 1.09x on the libritts\cite{18} dataset to 1.87x on the  Kitti dataset.

\textbf{Machine 2}. In Machine Environment 2, VersaDB only failed to maintain superior performance in the The Pile dataset. Below, we will discuss the performance improvements observed in datasets other than The Pile. In the Pure reading task, the speedup of VersaDB compared to Lmdb and TFRecord ranges from 1.1x on the ImageNet\cite{15} dataset to 2.49x on the  Youtube8M dataset. In the Shuffle reading task, VersaDB shows a speedup from 1.09x on the Squad dataset to 5.35x on the VOC dataset. In the Preprocessing task, the speedup of VersaDB ranges from 1.06x on the Kitti\cite{13} dataset to 4.12x on the Ljspeech\cite{19} dataset.

This progressive improvement in performance advantages
across datasets of increasing complexity demonstrates VersaDB’s particular strength in handling sophisticated data
types. The results suggest that VersaDB’s architecture
is especially well-suited for managing complex, multimodal
datasets, where its efficient data organization and processing
mechanisms provide increasingly significant benefits as data
complexity grows.

\begin{figure*}[t]

\centering
\subfloat[]{\includegraphics[width=1\textwidth, height=0.325\textheight]{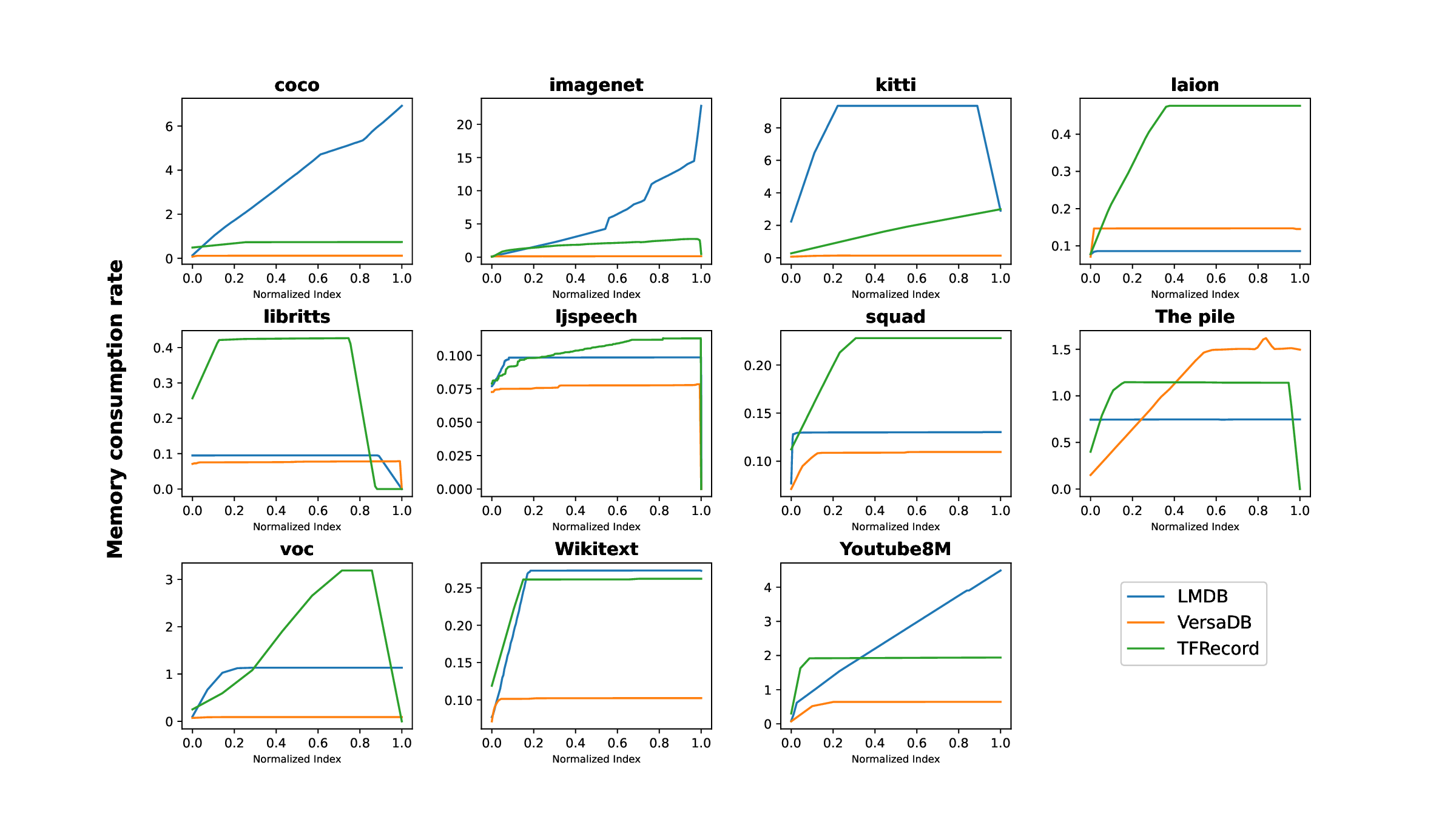}} 
\hfil
\subfloat[]{\includegraphics[width=1\textwidth, height=0.325\textheight]{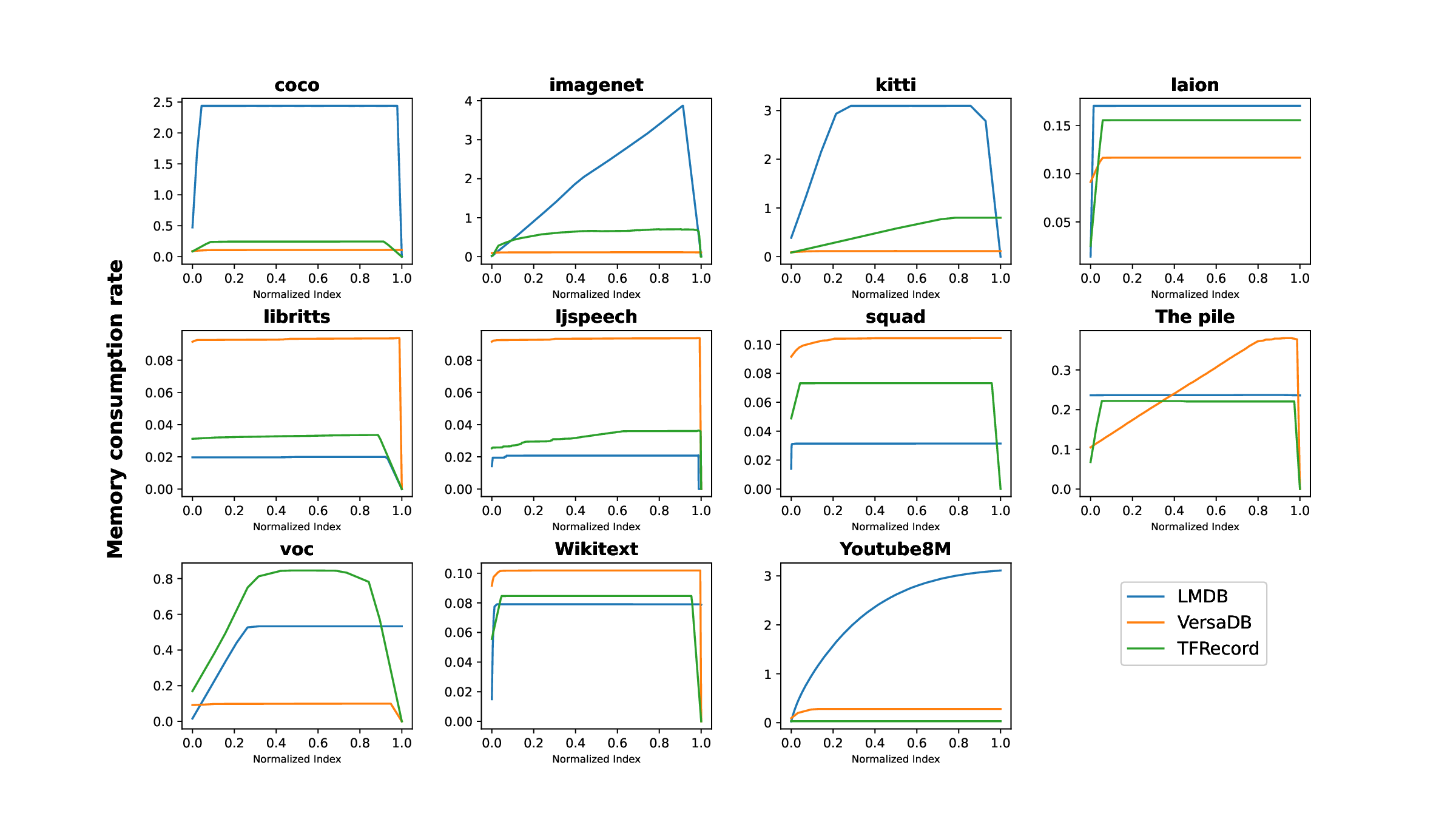} } 
\caption{Figures (a) and (b) represent the memory consumption rates of LMDB, VersaDB, and TFRecord across the eleven datasets in machine environments 1 and 2, respectively. The time consumed by the three methods is scaled to a unified scale. The x-axis represents the time interval, and the y-axis represents the memory consumption ratio, with consistent units of measurement.}
\label{fig:common source}
\end{figure*}

\begin{figure*}[t]

\centering
\subfloat[]{\includegraphics[width=0.8\textwidth, height=0.3\textheight]{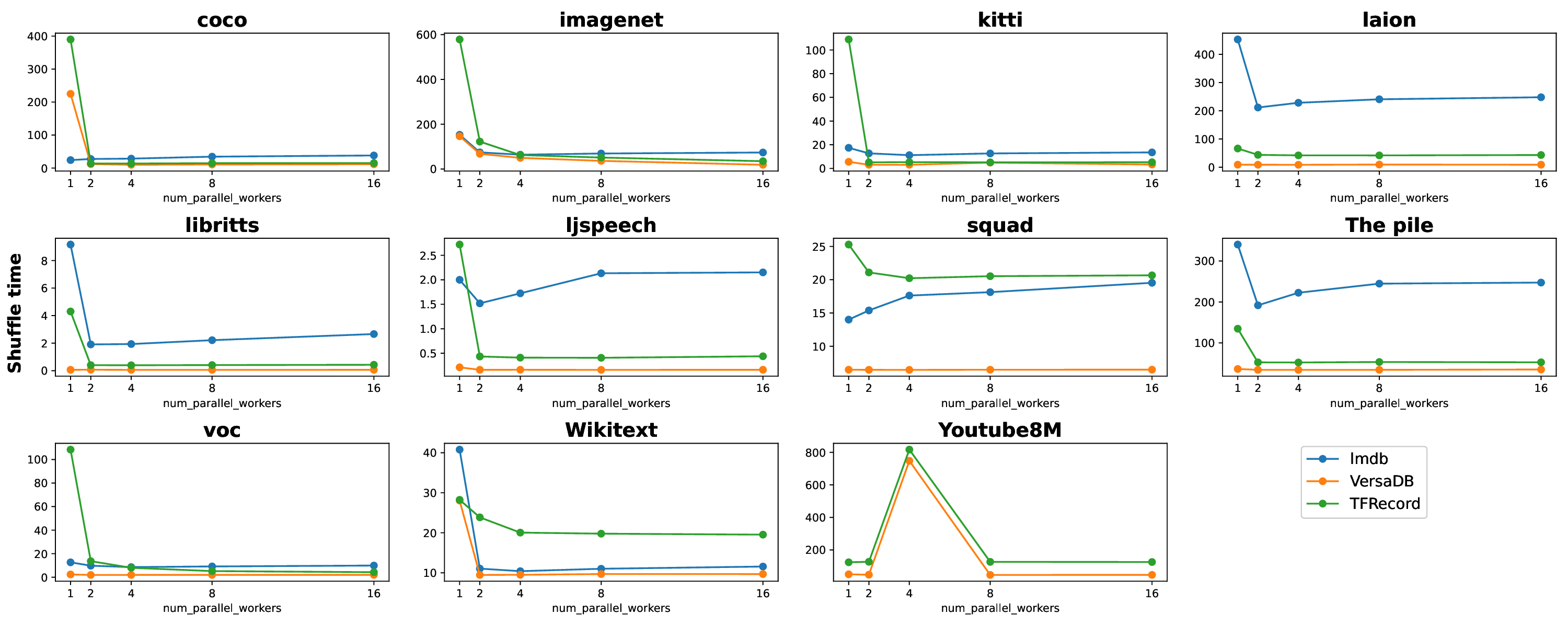}} 
\hfil
\subfloat[]{\includegraphics[width=0.8\textwidth, height=0.3\textheight]{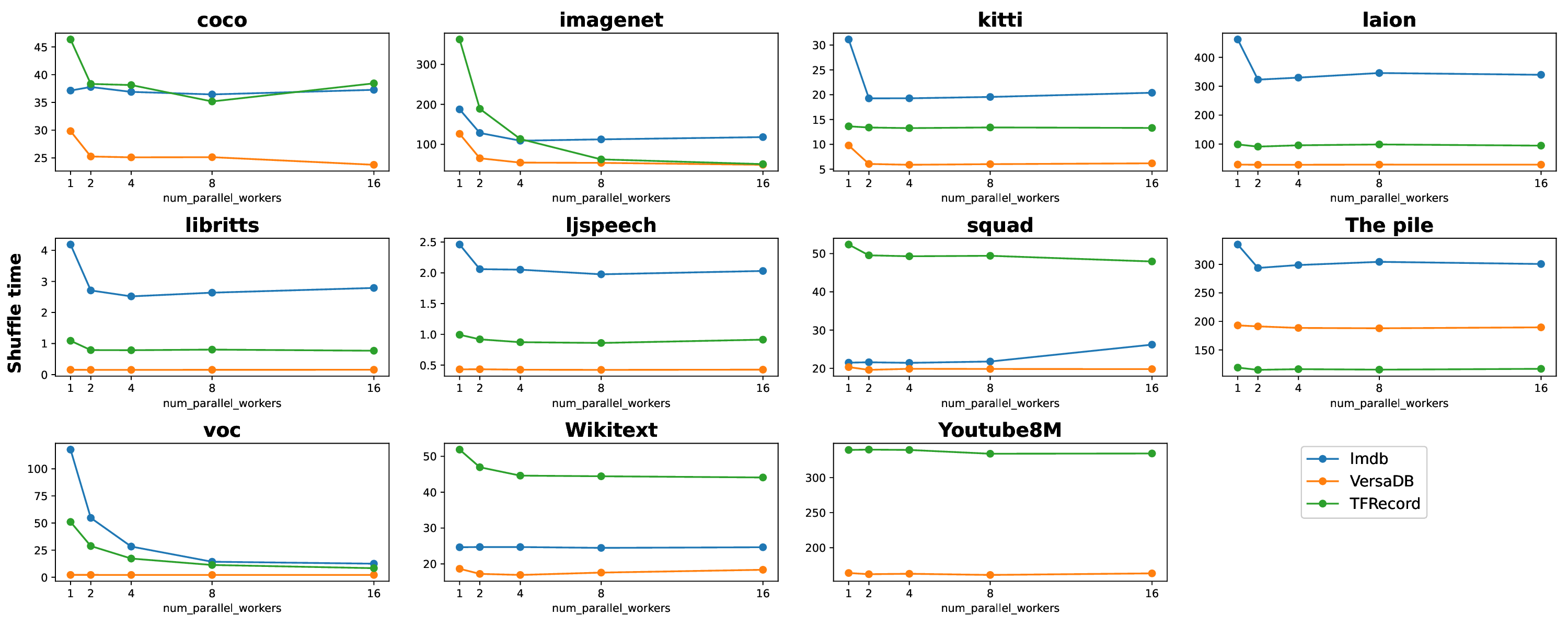} } 
\caption{Figures (a) and (b) show the test data for LMDB, VersaDB, and TFRecord on eleven datasets in machine environments 1 and 2, respectively. We set five parallelism levels: 1, 2, 4, 8, and 16. The x-axis represents the number of parallel workers, and the y-axis represents the consumption time.}
\label{fig:common worker}
\end{figure*}

\subsection{Comparison of Resource consumption}
In this experiment, we tested the resource usage of VersaDB, Lmdb, and TFRecord while performing the Shuffle reading task on eleven datasets. The tests were conducted in two machine environments, as shown in Fig. \ref{fig:common source}. 

\textbf{Machine 1}. In Machine Environment 1, VersaDB demonstrated better memory management, with memory consumption reduction effects of 82.9\%, 93.02\%, 11.2\%, 21.81\%, 17.56\%, 91.2\%, 59.07\% and 68.66\%, respectively for coco, Imagenet, Kitti, Libritts, Ljspeech, Squad\cite{17}, VOC, WikiText, Youtube8M. In the Laion dataset, the peak memory consumption was only 30.6\% of TFRecord, but the acceleration compared to TFRecord was 4.4x. In The Pile dataset, although the acceleration compared to TFRecord was higher, the peak memory consumption reached 130\% of TFRecord.

\textbf{Machine 2}. In Machine Environment 2, VersaDB shows advantages in memory usage on the COCO, ImageNet, Kitti, The laion and VOC\cite{12} datasets, with memory reduction effects of 49.8\%, 81.05\%, 78.19\%, 22.39\% and 79.51\%, respectively. In The Pile dataset, similar to Machine Environment 1, our peak memory consumption was the highest among three methods, reaching 160\% of TFRecord.

\subsection{Sensitivity Analysis}
In this experiment, to test the sensitivity of our database, we conducted tests on eleven datasets across two machine environments, evaluating the performance of TFRecord, VersaDB, and Lmdb under different parallel reading conditions. We set the parallelism levels to 1, 2, 4, 8, and 16. For the Youtube8M dataset, we did not present the LMDB test results, as its performance significantly lags behind that of VersaDB and TFRecord, making it inconvenient to display together.The result is shown in Fig. \ref{fig:common worker}.

\textbf{Machine 1}.In Machine Environment 1, our VersaDB showed a competitive advantage in all five parallel environments tested, with lower times compared to the other two methods. It maintained high acceleration performance across all parallel environments in LibriTTS, Ljspeech, SQuAD, VOC, Laion, Youtube8M and The Pile. However, in some other datasets, such as Imagenet, the acceleration effect of VersaDB became nearly identical to that of TFRecord in certain parallel environments, showing a decrease in acceleration performance.

\textbf{Machine 2}. In Machine Environment 2, our VersaDB demonstrated a competitive advantage in all five parallel environments across all datasets except The Pile, with shorter run times compared to the other two methods. Moreover, it sustained high acceleration performance across all parallel environments in all datasets except for Imagenet and The Pile. In the Imagenet dataset, the time consumption of VersaDB and TFRecord gradually converged as the parallelism level increased.

We believe that the reason VersaDB performs better in different parallel environments is due to VersaDB's architecture design, which minimizes thread contention through its page-based data organization and independent worker scheduling mechanism.

\subsection{Data parallel reading}
In this experiment, we tested the performance of VersaDB, TFRecord, and Lmdb in reading the Kitti, LJSpeech, and Wikitext datasets under different parallel distributed environments. These three datasets represent three common modalities. We used data parallelism for distributed reading, and tested the performance when the number of nodes was 2, 4, and 8. The number of shards in VersaDB was kept consistent with the number of nodes, meaning that when there were 2 nodes, there were 2 shards. Our testing environment consisted of a GPU cluster with eight NVIDIA 3090 GPUs.

Table III show that our database can fully leverage the advantages of multiple nodes, achieving up to 2.3x and 5.3x speedup on the Wikitext and Kitti datasets, respectively. In the LJSpeech dataset, we achieved a 40x speedup when the number of nodes was 8.

\setlength{\tabcolsep}{6pt} 
\begin{table}[t]
\centering

\begin{tabular}{|c|c|c|c|c|}
\hline
                           &  & \multicolumn{3}{c|}{Distributed reading} \\ \cline{2-5}
                          & Shard num & VersaDB & TFRecord  & Lmdb   \\ \hline
\multirow{3}{*}{Kitti} & 2 & \textbf{4.62}   & 8.03 & 9.9  \\ \cline{2-5}
                          & 4 & \textbf{1.15}   & 6.4 & 6.13   \\   \cline{2-5}
                          & 8 & \textbf{1.04}   & 6.08 & 4.34   \\ \hline
\multirow{3}{*}{Ljspeech} & 2 & \textbf{0.1}  & 2.09 & 1.24  \\ \cline{2-5}
                          & 4 & \textbf{0.051}  & 1.25 & 1.16  \\ \cline{2-5}
                          & 8 & \textbf{0.036}   & 1.6 & 1.45   \\ \hline
\multirow{3}{*}{Wikitext} & 2 & \textbf{5.43}  & 68.6 & 8.57  \\ \cline{2-5}
                          & 4 & \textbf{2.04}  & 34 & 4.66  \\ \cline{2-5}
                          & 8 & \textbf{1.26}   & 63.5 & 2.67   \\ \hline
\end{tabular}
\vspace{10pt}  
\caption{Performance comparison under different number of distributed nodes}
\end{table}
\section{Discussion}
\subsection{Fast indexing data}
The experimental results demonstrate VersaDB’s significant performance advantages, particularly in handling large-scale and complex datasets. Through
detailed analysis, we can attribute these performance improvements to several key architectural innovations in our design.

First, VersaDB’s page management system with scalarblob separation proves to be a crucial performance enabler. By segregating scalar data (e.g., labels, metadata) from blob data
(e.g., images, audio), the system achieves optimized storage
and retrieval strategies for different data types.

Second, our hybrid indexing mechanism plays a pivotal
role in achieving superior performance while LMDB relies
on B+ tree structures. VersaDB implements an adaptive indexing strategy that dynamically optimizes based on data characteristics and access.  This design proves particularly effective in scenarios
involving random access patterns
patterns.

Third, our schema-based data organization system enables efficient handling of complex data structures. Unlike
TFRecord’s fixed serialization approach or LMDB’s key-value
paradigm, VersaDB’s flexible schema allows optimized representation and access patterns tailored to specific data types.
\subsection{Resource management}
Our resource utilization
experiments reveal significant efficiency advantages of VersaDB in memory usage at most of time. These efficiency gains can be attributed
to several key architectural designs that specifically target resource optimization. The most notable improvement comes from our innovative
page management system, which demonstrates remarkable
memory efficiency.

First,  our scalar-blob separation strategy, which enables selective loading of data components.

Second, our dynamic page replacement policy,
which efficiently manages memory allocation and deallocation
based on actual usage patterns

These resource efficiency improvements directly address
critical challenges in modern AI workflows. The significant
reduction in memory usage enables processing of larger datasets on existing infrastructure, reducing hardware requirements and associated costs.

\section{conclusion}
In this paper, we tested various datasets include NLP, CV, multimodal, and audio datasets. VersaDB consistently maintains a performance advantage in Pure Reading, Shuffle Reading, and Preprocessing tasks under default settings in two common machine environments, with acceleration ranging from 1.06x to 5.35x, except for The Pile dataset tested in Machine Environment 2. In Machine Environment 1, VersaDB has lower memory usage compared to LMDB and TFRecord at most of time, but in Machine Environment 2, it exhibits higher memory consumption for audio and NLP datasets. We also tested performance across different parallelism levels, and VersaDB showed performance advantages at all parallelism levels in both machine environments except for The Pile dataset tested in Machine Environment 2. We also tested the reading performance under different distributed parallel environments on three common modality datasets, and VersaDB consistently maintained a performance advantage in all environments.

In conclusion, VersaDB demonstrates strong performance and resource management across different categories of datasets, maintaining a good adaptability to various machine environments and architectures.

\section{Limitation and future work}
In terms of memory consumption, we found that on the AARCH64 architecture, VersaDB consumes more memory than LMDB and TFRecord in NLP and audio datasets, while on the x86-64 architecture, this occurs only in The Pile dataset. 

In terms of performance, after testing, the performance of VersaDB on The Pile dataset in the AARCH64 architecture was not ideal. Therefore, in the future, we will focus on further optimizations for the AARCH64 architecture to achieve consistent high performance across mainstream machine architectures.

In terms of usability, we will continue to track popular datasets and provide APIs for converting these datasets to the VersaDB format. At the same time, we will also consider the compatibility of conversions between different AI data formats. For example, the conversion from TFRecord to VersaDB currently does not support converting TFRecord's variable data types.
\bibliographystyle{unsrt}  
\bibliography{referance}

\end{document}